%% file: main.tex
\documentclass[letterpaper]{article} 
\usepackage[]{aaai25}  
\usepackage{times}  
\usepackage{helvet}  
\usepackage{courier}  
\usepackage[hyphens]{url}  
\usepackage{graphicx} 
\usepackage{natbib}  
\usepackage{caption} 
\usepackage{graphicx}
\usepackage{subcaption}
\usepackage{makecell}
\usepackage{algorithmic}

\usepackage{amsmath}
\usepackage[linesnumbered,ruled]{algorithm2e}

\usepackage{newfloat}
\usepackage{listings}
\DeclareCaptionStyle{ruled}{labelfont=normalfont,labelsep=colon,strut=off} 
\title{Rack Position Optimization in Large-Scale Heterogeneous Data Centers}

\author {Chang-Lin Chen\textsuperscript{\rm 1}, Jiayu Chen\textsuperscript{\rm 1}, Tian Lan\textsuperscript{\rm 2}, Zhaoxia Zhao\textsuperscript{\rm 3}, Hongbo Dong\textsuperscript{\rm 3}, and Vaneet Aggarwal\textsuperscript{\rm 1}
}
\affiliations {
    \textsuperscript{\rm 1}Purdue University\hspace{.3in}
        \textsuperscript{\rm 2}George Washington University\hspace{.3in}
         \textsuperscript{\rm 3}Meta Platforms, Inc. 
}

\usepackage{bibentry}

\begin{document}

\maketitle

\begin{abstract}

As rapidly growing AI computational demands accelerate the need for new hardware installation and maintenance, this work explores optimal data center resource management by balancing operational efficiency with fault tolerance through strategic rack positioning considering diverse resources and locations.
Traditional mixed-integer programming (MIP) approaches often struggle with scalability, while heuristic methods may result in significant sub-optimality. To address these issues, this paper presents a novel two-tier optimization framework using a high-level deep reinforcement learning (DRL) model to guide a low-level gradient-based heuristic for local search. The high-level DRL agent employs Leader Reward for optimal rack type ordering, and the low-level heuristic efficiently maps racks to positions, minimizing movement counts and ensuring fault-tolerant resource distribution. This approach allows scalability to over 100,000 positions and 100 rack types. Our method outperformed the gradient-based heuristic by 7\% on average and the MIP solver by over 30\% in objective value. It achieved a 100\% success rate versus MIP's 97.5\% (within a 20-minute limit), completing in just 2 minutes compared to MIP's 1630 minutes (i.e., almost 4 orders of magnitude improvement). Unlike the MIP solver, which showed performance variability under time constraints and high penalties, our algorithm consistently delivered stable, efficient results—an essential feature for large-scale data center management.

\end{abstract}

\input{1introduction}

\input{2background}
\input{3system}
\input{4problem}
\input{5proposed_framework_alg}

\input{6simulation}

\input{7conclusion}
\bibliography{arXiv/aaai25}
\clearpage
\appendix
\input{apdx}

\input{apdx_sims}
\end{document}

%% file: 1introduction.tex
\section{Introduction}



In large-scale data centers, efficient rack movement and resource allocation are important for balancing multiple operational objectives with respect to resource-efficiency, fault tolerance, and sustainability goals. 
Rising data center investments by major tech companies \cite{meta_invest, microsoft_invest} and increasing demands for data center efficiency highlight the importance of optimized rack positions, which directly impact movement counts, operational efficiency, and overall system resilience \cite{abbas2021}. 
As data centers grow in complexity, continuous hardware updates and resource reallocation are needed to meet evolving service requirements. Optimized rack placement reduces movement counts, simplifies load balancing, and enhances fault tolerance—an increasingly important factor as data centers host diverse, high-performance hardware with varied resource needs. Scalable and adaptable optimization solutions are thus important to sustain service quality, operational resilience, and environmental efficiency in modern data center environments, which is the focus of this paper.

Previous works have demonstrated the substantial impact of rack position optimization on data center performance. Seminal research by \cite{datacenter_energy_Luiz} established how rack layout designs significantly reduce operational energy usage, while \cite{George2022_intra_rack_disaggr} proved through production HPC experiments that rack-level resource disaggregation markedly improves utilization. Thermal management advances were made by \cite{ZHANG2023_cool}, 
demonstrating how optimized rack layouts enhance heat dissipation and energy efficiency. For heterogeneous environments, \cite{rack_layout_design_heter_dc} developed effective rack-position mapping techniques for optimal cooling performance. Resource management innovations include \cite{racksched}'s microsecond-scale scheduler and \cite{rackmem}'s performance-boosting caching layer for rack-scale computing, collectively establishing rack positioning as important for both energy and resource optimization.



Composable data centers provide a promising solution by disaggregating resources like CPU, memory, and storage, enabling dynamic pooling and reallocation based on workload requirements \cite{composable_arch_Li_2016}. This architectural approach stands in stark contrast to traditional systems that synchronize hardware replacement, which often leads to inefficiencies and delayed adoption of new technologies. The flexibility offered by composable architectures significantly enhances resource utilization and operational efficiency, making them particularly valuable for fast-evolving, heterogeneous AI environments. This architectural flexibility also removes many traditional constraints on rack positioning, enabling more opportunities for optimizing physical infrastructure layout. Strategic rack relocation simplifies scheduling, balances load distribution, and reduces maintenance overhead, especially as heterogeneity increases in data centers \cite{datacenter_energy_Luiz,regaieg_multi-objective_2021}. Furthermore, the ongoing need to install new racks to meet growing demand underscores the importance of optimizing rack-position mapping at the data center level to maintain operational efficiency. Regular maintenance of building infrastructure, power devices, and other facilities further necessitates the ability to efficiently move and position racks, ensuring minimal disruption during these routine activities.

This paper considers a region of rack-level disaggregated data centers consisting of heterogeneous racks and positions.
To the best of our knowledge, this is the first work to optimize rack positioning for fault tolerance, ensuring resources from a subset of rack types are well-distributed across various scopes. Such fault tolerance empowers the hardware layer to provide reliability and availability guarantees to software services, important for optimal resource usage in large organizations. Consequently, we aim to identify a rack-position mapping that optimizes a weighted combination of rack movement count and fault tolerance metrics, while adhering to multi-layer resource constraints, including energy consumption, demand fulfillment, and placement limits. This novel problem is formulated as an integer non-linear optimization problem (INLP), for which computationally efficient approaches are lacking.
To solve this problem, this paper proposes a two-tier optimization approach, with the Leader Reward approach \cite{wang2024leaderreward} guiding the higher-level ordering of rack-type subproblems, and a gradient-based heuristic algorithm addressing each subproblem at the lower level, enabling efficient decomposition and targeted optimization for each rack type. Leader Reward, designed for standard combinatorial problems like the traveling salesman problem (TSP) and capacitated vehicle routing problem (CVRP), requires substantial memory due to its transformer-based architecture, making it impractical for direct application to our large-scale problem. Further, our problem does not have similar network/graph structure as in TSP/CVRP limiting direct usage of the approach. Therefore, we adapt it to our unique problem structure by incorporating an additional encoder, problem decomposition, and a tailored heuristic. The approach is shown to be scalable and outperform the MIP solvers as well as the greedy heuristic alone.

{\bf Summary of Contributions: } The main contributions in this work are summarized as follows:

\noindent {\bf 1. } We introduce a rack movement optimization framework tailored for composable, heterogeneous data centers, which minimizes combined metrics associated with rack movement and fault tolerance. The problem is motivated by practical needs and can significantly advance datacenter intelligent management.


\noindent {\bf 2. } We formulate the problem as a computationally challenging INLP and address scalability issues by integrating a deep reinforcement learning (DRL) algorithm with a gradient-based heuristic in a tiered, collaborative architecture. The DRL algorithm employs Leader Reward, a technique well-suited for combinatorial tasks, to optimally order rack types, while the heuristic efficiently assigns specific positions. Experimental results demonstrate the heuristic's competitiveness and speed.


\noindent {\bf 3. } Simulation results demonstrate the proposed framework’s superior performance, achieving a 7\% improvement over a heuristic-only approach. Further, the proposed approach outperforms the MIP solver in terms of both the execution time and objective value.


%% file: 2background.tex
\section{Related Works}

Data center optimization has evolved through diverse methods, including ILP, MILP, heuristics, and more recent advancements in DRL, each targeting various challenges from power management to infrastructure optimization. However, rack layout optimization in disaggregated cloud systems remains largely unexplored. 
\paragraph{Data center optimization methods}
Research on data center optimization spans several approaches. For power optimization, \cite{pahlavan_power_2014} uses Integer Linear Programming (ILP) with heuristics to minimize active chassis while maintaining performance, while \cite{engy_workload_pl_comp_dc_2021_ajibola} applies MILP and greedy algorithms for workload placement in composable architectures. For VM management, \cite{stefanello2019hybrid} combines MILP with genetic algorithms across geo-distributed data centers, and \cite{Xia2017} uses hierarchical decomposition with MIP to address disk anti-colocation constraints.
For infrastructure management, the Radar framework \cite{radar2023} tackles cooling challenges using MILP and heuristics, while \cite{flow_table_xue_2021} optimizes flow table installation in PDP switches using MINLP and k-median algorithms. Recent DRL applications include fault tolerance by logical cluster formation \cite{ras2024},
and network-aware resource allocation using GNNs \cite{drl_networkaware_gnn,drl_mix_flow_sche}. However, optimizing rack layouts for new rack installations and maintenance, while accounting for broader management requirements such as fault tolerance, cooling, and energy efficiency, remains an unexplored area.

\paragraph{DRL methods for combinatorial problems}
\cite{kool2019attention} introduced attention-based architectures for neural combinatorial optimization (NCO), specifically tackling routing problems like TSP and VRP, which eliminated the need for handcrafted heuristics and established attention mechanisms as foundational in NCO. \cite{pomo} expanded on this with POMO, a multi-optima framework that optimizes across multiple optimal solutions within each instance, significantly enhancing convergence speed and solution diversity. \cite{grinsztajn2023poppy} then developed Poppy, a population-based reinforcement learning approach where a set of agents explores diverse solution paths, promoting thorough exploration and mitigating premature convergence. \cite{chalumeau2024compass} introduced Compass, which employs latent space search to dynamically adapt policies in response to problem structure variability, enabling the model to handle increasingly complex combinatorial tasks. Finally, \cite{wang2024leaderreward} extended POMO by introducing a leader reward mechanism that prioritizes solutions near optimal regions, enhancing stability and precision by emphasizing high-quality solutions. Together, these methods illustrate substantial advancements in NCO through attention-driven models, policy diversity, adaptive exploration, and enhanced reward structures tailored to complex optimization problems.


Directly applying existing approaches is infeasible for our problem due to its large scale and unique structure, which differs from standard combinatorial problems like TSP and CVRP. To address this, we integrate the Leader Reward approach \cite{wang2024leaderreward} with a gradient-based heuristic, selecting it over Compass and Poppy for its superior performance and memory efficiency \cite{chalumeau2024compass}—important for large-scale applications. Further, our algorithm can be directly used for online optimization for dynamic system environments since it captures the dynamic movement count.


%% file: 3system.tex
\section{System Model}\label{sec:system_model}

This paper explores the strategic placement and adjustment of racks within data centers to address user needs and system requirements, and ensure fault-tolerant resource allocation. Users submit requests for a range of resources, such as CPUs, memory, disk storage, GPUs, and more. Consequently, the placement and position adjustment of racks is meticulously determined based on the evolving demands for these diverse resource types. Each rack is categorized into a specific type according to the combination of resources it offers. For example, racks equipped with GPUs and memory form one category, whereas those outfitted with CPUs and disk storage constitute another.

Within the complex ecosystem of data centers, designated spots known as positions define where each rack can be located, with the stipulation that each position can accommodate only one rack. These positions are aggregated into sets known as scopes, which are physical clusters such as regions, individual data centers, suites, and main switchboards (MSB). Some of the scopes are organized in a hierarchical fashion, where larger entities encompass several smaller ones, such as a region containing multiple data centers. The primary goal is to minimize system metrics, which include rack movement counts and 
penalties for resource imbalances across scopes, while simultaneously responding adaptively to user demands. This proactive rack repositioning is designed to enhance the system’s overall responsiveness and operational efficiency.

\begin{figure}[t]
    \centering
    \includegraphics[scale=0.3]{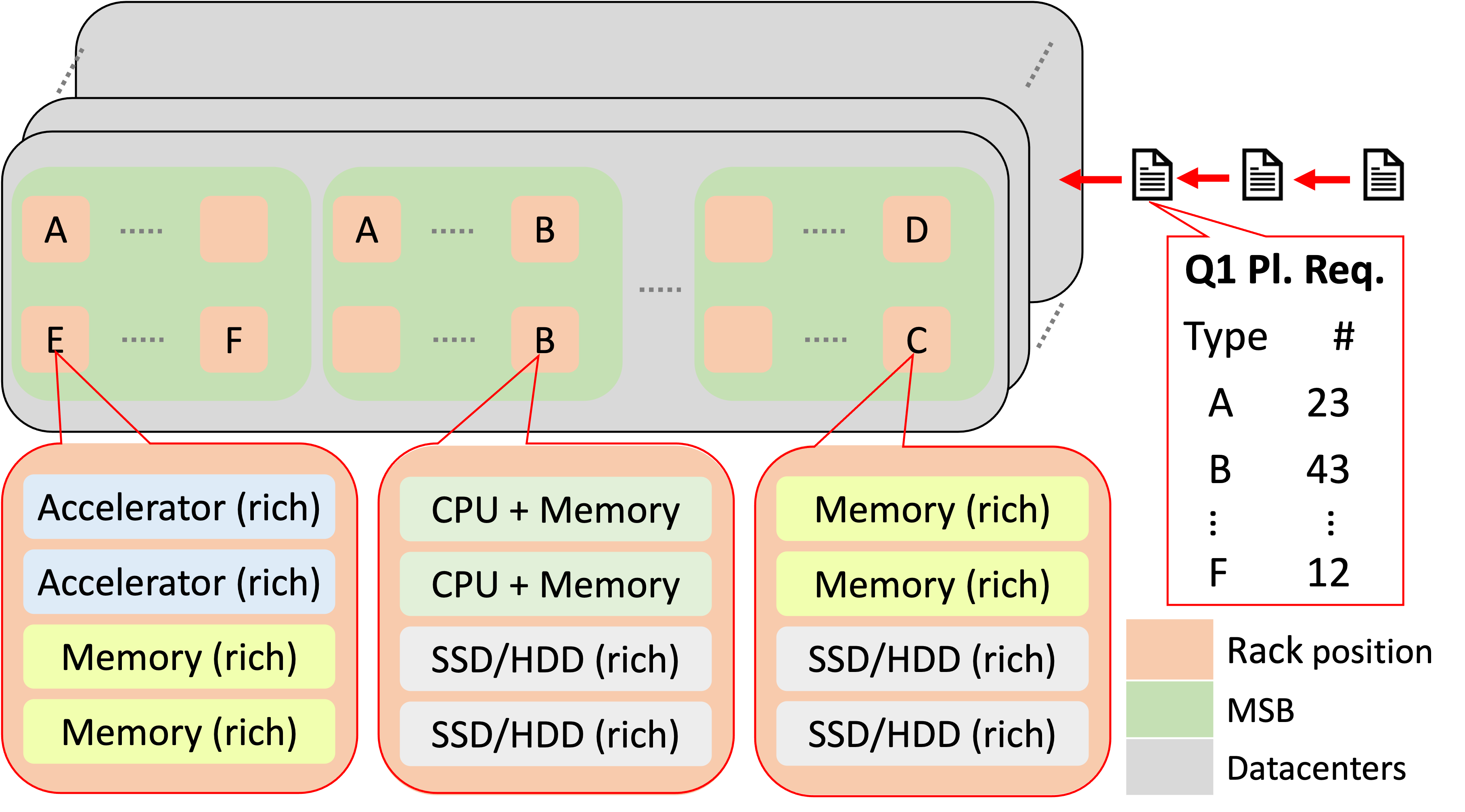}
    \caption{Simplified depiction of a rack movement system in heterogeneous data centers. The figure illustrates different rack types, labeled A, B, C, D, E, and F, each consisting of server nodes rich in specific resource types. The color-coded legend at the bottom represents different scopes within the datacenters, each with its resource limits. The figure also highlights placement requests, which indicate the number of racks required to meet resource usage demands. The rack movement system satisfies placement requests and service level objectives (SLOs) by dynamically moving racks or placing new racks within the data centers.}
    \label{fig:rackmovement}
\end{figure}

Let $\mathbf{K}$ be the set of rack types. In this model, we assume that any two racks of the same type are fungible, so we do not work with individual rack. Let $\mathcal{G}$ be a collection of subsets of $\mathbf{K}$, which we use to define metrics and constraints. Each $G \in \mathcal{G}$ may represent racks used for similar purposes, such as storage, computation, AI training, or racks used by specific product teams.
Let $\mathbf{P}$ be the set of all possible positions for hosting racks (expected to be on the order of $10^5$), and let $\mathcal{S}$ be a collection of subsets of $\mathbf{P}$. Each $s \in \mathcal{S}$ represents a position scope, such as a data center, MSB, or suite. In later formulations, we use $S$ to represent a Boolean matrix where $S_{ij} = 1$ if position $i$ is included in scope $j$.
Let $R$ be a resource matrix in which each row represents a rack type (indexed as in $\mathbf{K}$), and each column represents a resource type (e.g., power/cooling consumption, CPU, storage) or a Boolean flag (e.g., whether it is an AI rack).
Define $X = [x_{p,k}]{p \in \mathbf{P}, k \in \mathbf{K}}$ as a Boolean matrix of size $|\mathbf{P}| \times |\mathbf{K}|$, where the element $x{p,k}$ at row $p$ and column $k$ is a binary variable indicating whether rack type $k$ is present at position $p$. Position $p$ is vacant if all entries in row $p$ of $X$ are zero, i.e., $x_{p,k} = 0$ for all $k$. The placement request for rack type $k$ is denoted by $d_k$. To regulate resource overuse, a placement limit $q$ is introduced to control the number of new racks placed in the system.
\paragraph{Position Mapping Constraint}
Assuming a position can accommodate at most one rack at a time, position-to-rack-type mapping, $X$, should satisfy the following constraint,
\begin{equation} \label{eq: pos_rack_mapping_constr}
    g_{1,p}(X) = \sum_{k\in \mathbf{K}}x_{p,k} \leq 1, \forall p \in \mathbf{P}.
\end{equation}
\paragraph{Demand Fulfillment}
Moreover, to guarantee that the supply of rack types is enough to satisfy the placement request, $X$ should satisfy the following constraint,
\begin{equation}\label{eq: demand fulfillment}
    g_{2,k}(X)=\sum_{p\in\mathbf{P}}x_{p,k}-d_{k}\geq 0, \forall k \in \mathbf{K}.
\end{equation}
\paragraph{Placement Limit}
The total number of racks newly placed should be less than the placement limit $q$.
\begin{equation}\label{eq: placement_limit}
    g_{3}(X)=\sum_{k\in\mathbf{K}}\max\Big(0,\sum_{p\in\mathbf{P}}x_{p,k}-\sum_{p\in\mathbf{P}}\bar{x}_{p,k}\Big)-q\leq 0,
\end{equation}
where $\bar{X} = [\bar{x}_{p,k}]{p \in \mathbf{P}, k \in \mathbf{K}}$ represent the previous rack-position mapping. The number of newly placed racks are calculated by comparing the number of racks placed at current and previous time slots for all rack types and summing all the positive differences.

We aim to minimize rack movement and optimize resource distribution of all the resource types across various scopes. 
\paragraph{Rack Movement Count}
When a rack moves to another position, it makes resources unavailable to the users and induces penalty of moving racks physically.  
The movement count of rack type $k$ at position $p$ is given as
\begin{equation}
    o_{1,p,k}(X) = M_k\times\max(0, \bar{x}_{p,k}-x_{p,k}),
\end{equation}
where $M_{k}$ is weight of the movement count of rack type $k$. 
\paragraph{Resource Spread Metrics}
For fault tolerance, resources from a subset of rack types should be well-distributed across a group of scopes to ensure resource availability and reliability in the event of a failure within any physical scope. Given the set of resource spread requirement ${r_i, G_i, \mathbf{c}_i}$, where $r_i$ is a specific resource type, $G_i\subset \mathbf{K}$ is a subset of rack types, and $\mathbf{c}_i$ is a union of mutually-exclusive scopes, the spread metrics is given as
\begin{equation}\label{eq: resource_spread}
    o_{2,i}(X) = \tau(S[:,\mathbf{c}_i]^T\cdot X[\mathbf{c}_i,G_i]\cdot R[G_i,r_i]),
\end{equation}
where $S[:,\mathbf{c}_i]^{T}\cdot X[\mathbf{c}_i,G_i]\cdot R[G_i,r_i]$ is the resource utilization of resource type $r_i$ derived from rack types in $G_i$ across the scopes in $\mathbf{c}_i$ and $\tau$ is a multivariate function to measure the spread such as max, standard deviation, etc.

\paragraph{Scope Resource Consumption Limit}
To meet management requirements, such as limits for CPU allocation, energy consumption and bandwidth usage, we introduce scope-specific resource constraints for each resource type and scope. Let $L$ represent the resource limit matrix of size $|\mathcal{S}|\times |\mathbf{R}|$, where each element $L[s,r]$ denotes the resource limit for resource type $r$ in scope $s$. The resource limit constraint for each scope and resource type pair is then defined as:
\begin{equation}\label{eq: resource_limit_constr}
    o_{3,s,r}(X) = \zeta((L-S^T\cdot X \cdot R)[s,r]), \forall s \in \mathcal{S}, r \in \mathbf{R}.
\end{equation} 
where $\zeta(\cdot)$ is a function that returns a positive value for positive inputs and zero for non-positive inputs.

We summarize the notions in Table \ref{table: notation_summary}. The following section shows the optimization problem considered and how it is optimized.

%% file: 4problem.tex
\section{Problem Formulation}\label{sec:problem_formulation}
This section presents the problem considered in this work.
The utility function is formulated considering the rack movement count and resource spread across scopes as,
\begin{equation} \label{eq:total_utility}
    \mathcal{U}=\sum_{k\in \mathbf{K}}\sum_{p\in \mathbf{P}}o_{1,p,k}+\beta_1\sum_{i=1}^{|I|}o_{2,i}+\beta_2\sum_{s\in\mathcal{S}, r\in\mathbf{R}}o_{3,s,r},
\end{equation}
where $\beta_1$ and $\beta_2$ are the weights for the resource spread and resource limit penalties, respectively.

In this work, we aim to obtain high-quality solutions while considering the constraints of position mapping, demand fulfillment, and placement limit.
We define our optimization problem and present it as:
\begin{align}\label{eq:optimization_problem}
    \min_{x_{p,k},\forall p\in \mathbf{P}, k \in \mathbf{K}} &\mathcal{U}(X)\\
    s.t.\quad\quad &g_{1,p}(X) \leq 1, \forall p \in \mathbf{P},\\
    &g_{2,k}(X) \geq 0, \forall k \in \mathbf{K},\\
    &g_{3}(X)\leq 0,
\end{align}\label{prob:original_problem}
The above problem is a integer programming problem with the number of variables being $|\mathbf{P}||\mathbf{K}|$ and the number of constraints being $|\mathbf{P}|+|\mathbf{K}|+1$. 
We note that the number of variables can go beyond millions for a region of data centers with over 100k positions and 100 rack types, limiting using commercial solvers directly. 
To obtain optimal results efficiently, we design a scalable heuristic approach integrated with a DRL-based combinatorial solver. The following section details each component of our proposed algorithm.

%% file: 5proposed_framework_alg.tex
\section{Proposed Algorithm}\label{sec:proposed_algorithm}
In this section, we present our proposed DRL-based approach to solving the rack movement optimization problem.
Given the large scale of datacenters and the interdependencies between rack types as specified in functions \eqref{eq: placement_limit}, \eqref{eq: resource_spread}, and \eqref{eq: resource_limit_constr}, we decompose the original optimization problem into subproblems by rack type. Each subproblem focuses on rack positioning for a single rack type, thereby reducing the number of variables and simplifying the overall complexity.
The problem formulated in \eqref{eq:optimization_problem} can become a integer nonlinear programming problem (INLP) depending on the choice of functions $\tau(\cdot)$ in \eqref{eq: resource_spread} and $\zeta(\cdot)$ in \eqref{eq: resource_limit_constr}. As most existing solvers cannot handle such problems efficiently, we develop a gradient-based heuristic approach for the subproblems that can address any differentiable variant of the problem.
Our gradient-based heuristic calculates the gradient of a composite function combining the objective function and the penalty for constraint violations. The algorithm iteratively flips either empty positions or positions occupied by the target rack type until the number of racks matches the requested quantity. The rack-position mappings are updated sequentially by applying the gradient-based heuristic to each rack type, one at a time, from the first to the last.
The ordering of rack type optimization significantly impacts the solution quality due to varying characteristics of each rack type, including their resource compositions, current system quantities, and demand levels. For example, optimizing rack types with lower demand first provides greater flexibility for subsequent rack types to achieve better objective values while satisfying constraints. Therefore, we employ a DRL-based combinatorial problem solver to optimize the sequence of suboptimizations.
The complete algorithm comprises a gradient-based heuristic for rack-position mapping and a DRL-based solver for sequencing suboptimizations. We detail the subproblems in Section \ref{subsec: subproblem}, followed by the gradient-based heuristic and DRL-based solver in Sections \ref{subsec: gradient_based_heuristic} and \ref{subsec: proposed_algorithm}, respectively.

\subsection{Suboptimization} \label{subsec: subproblem}
The original optimization problem in \eqref{eq:optimization_problem} is decomposed into subproblems based on rack types:
\begin{align}\label{eq:sub_optimization_problem}
    \min_{x_{p,k},\forall p\in \mathbf{P}} \;&\mathcal{U}(X^{\bar{\mathbf{K}}})\\
    s.t.\;\quad &g_{1,p}(X^{\bar{\mathbf{K}}}) \leq 1, \forall p \in \mathbf{P}, \\
    &g_{2,k}(X^{\bar{\mathbf{K}}}) \geq 0,\\ 
    &g_{3}(X^{\bar{\mathbf{K}}})\leq 0,
\end{align}\label{prob:sub_original_problem}
where $k \notin \bar{\mathbf{K}}$ denotes the rack type for the current subproblem, $X^{\bar{\mathbf{K}}}$ represents the rack-position mapping matrix with rack types in $\bar{\mathbf{K}}$ determined by previous subproblem solutions, and constraints $g_{1,p}$ and $g_{3}$ follow the same structure as in \eqref{eq:optimization_problem} but operate on $X^{\bar{\mathbf{K}}}$.

\subsection{Gradient Based Heuristic Algorithm} \label{subsec: gradient_based_heuristic}
We propose a gradient-based heuristic algorithm that sequentially solves the subproblem in \eqref{eq:sub_optimization_problem} for each rack type to obtain a solution to the original optimization problem in \eqref{eq:optimization_problem}. The algorithm iteratively updates rack positions by following the gradient of an augmented objective function:
\begin{equation}\label{eq: gradient_objective}
f(X^{\bar{\mathbf{K}}}; k) = \mathcal{U}(X^{\bar{\mathbf{K}}})+\gamma g_{3}(X^{\bar{\mathbf{K}}}),
\end{equation}
where $k$ denotes the rack type being optimized (corresponding to the variable column in $X^{\bar{\mathbf{K}}}$), $g_{3}$ is the inequality constraint from \eqref{eq:sub_optimization_problem}, and $\gamma > 0$ is the penalty coefficient for constraint violations. For each rack type, the algorithm enforces constraints $g_{1,p,k}$ and $g_{2,k}$ by: (1) using a masking mechanism to exclude non-applicable positions (positions already assigned to other rack types when additional positions are needed, or vacant positions when some need to be cleared), and (2) ensuring the required number of racks $d_k$ for type $k$. Finally, $\Gamma$ rounds of adjustments are applied to further minimize the objectives.
The complete algorithm is presented in Algorithm \ref{alg: gradient_based_heuristic}.
\begin{algorithm}[t]
  \DontPrintSemicolon
  \SetAlgoLined
    \KwIn{\{$X$, $S$, $R$, $L$, $d_k$, $q$, $\{(\mathbf{c}_i, G_i, r_i)\}$, $\bar{\mathbf{K}}$, $\Gamma$\}}
    \KwOut{$\{x_{p,k}, \forall p\in \mathbf{P}\}$}
    \For{$k=1$ \KwTo $|\mathbf{K}|$}{
        $allocations \gets \sum_{p\in\mathbf{P}}x_{p,k}-d_k$\;
        $iter \gets abs(allocations)$\;
        \For{$i=1$ \KwTo $iter$}{
          \eIf{$allocations<0$}{
              $\Phi \gets \{p'| \sum_{k\in\mathbf{K}}x_{p,k}==0\}$\;
              $x_{p',k} \gets 1$, where $p'=argmax\nabla_{X^{\bar{\mathbf{K}}}}f(X^{\bar{\mathbf{K}}})\cdot \sum_{p\in\Phi}e_p$ and $e_p$ is an elementary vector\;
          }{
              $\Phi \gets \{p'| \sum_{k\in\mathbf{K}}x_{p,k}==1\}$\;
              $x_{p',k} \gets 0$, where $p'=argmax -\nabla_{X^{\bar{\mathbf{K}}}}f(X^{\bar{\mathbf{K}}})\cdot \sum_{p\in\Phi}e_p$\;
          }
        }
        \For{$i=1$ \KwTo $\Gamma$}{
            $\Phi \gets \{p'| \sum_{k\in\mathbf{K}}x_{p,k}==1\}$\;
            \If{$\Phi!=\phi$ (empty set)}{
                $x_{p',k} \gets 0$, where $p'=argmax -\nabla_{X^{\bar{\mathbf{K}}}}f(X^{\bar{\mathbf{K}}})\cdot \sum_{p\in\Phi}e_p$\;
                $x_{p',k} \gets 1$, where $p'=argmax\nabla_{X^{\bar{\mathbf{K}}}}f(X^{\bar{\mathbf{K}}})\cdot \sum_{p\in\Phi^c}e_p$ \;
            }
        }
    }
    \caption{Proposed heuristic algorithm\label{alg: gradient_based_heuristic}}
\end{algorithm}

\begin{table*}[t] 
\small
\caption{Parameter Setup} \label{table: 1}
\centering
\resizebox{\textwidth}{!}{%
 \begin{tabular}{|| c || c | c | c | c | c | c | c | c | c | c | c ||} 
 \hline
 Parameter & $|\mathbf{P}|$  & $|\mathbf{K}|$ & $|\mathbf{R}|$ & $|\mathcal{S}_1|$ & $|\mathcal{S}_2|$ & $|\mathcal{S}_3|$ & $[d_{min}, d_{max}]$ & $[q_{min}, q_{max}]$ & $[L_1^{max}, L_1^{min}]$ & $[L_2^{min}, L_2^{max}]$ & $[L_3^{min}, L_3^{max}]$\\
 \hline\hline
Value & 1000 & 10 & 10 & 2 & 10 & 50 & [20, 60] & [800, 1000] & [3,6] & [30,50] & [90, 110] \\
 \hline
 \end{tabular}
 }
\end{table*}

\subsection{Proposed Two-Tier Algorithm} \label{subsec: proposed_algorithm}
\begin{figure}[t]
    \centering
    \includegraphics[scale=0.3]{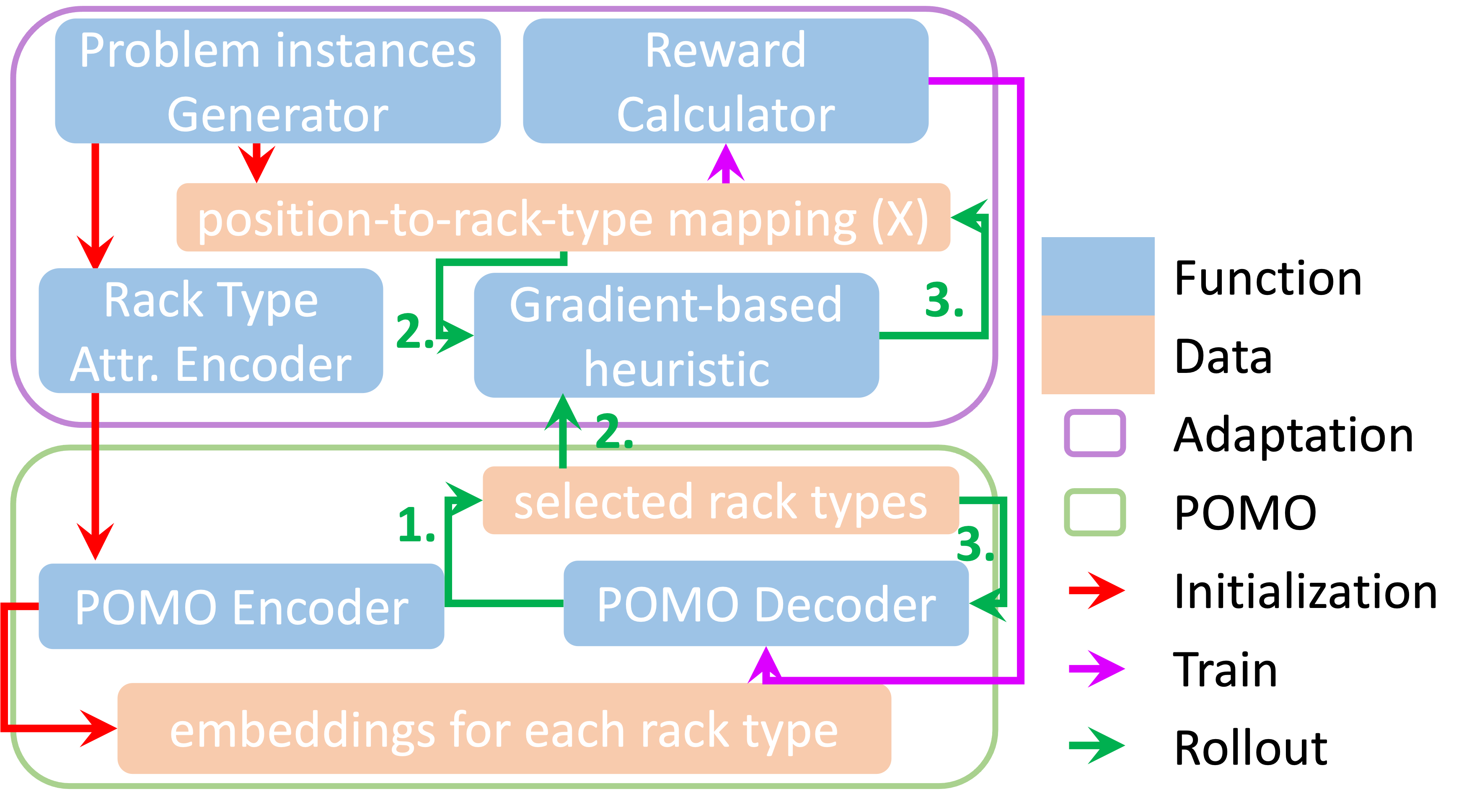}
    \caption{Our proposed algorithm integrates Leader Reward and adaptation components for rack positioning optimization. It comprises two main modules: an adaptation module (purple box) containing the problem generator and gradient-based heuristic; and a POMO module (green box) featuring an encoder-decoder architecture. The workflow encompasses initialization (red arrows), training (purple arrows), and rollout (green arrows) phases, with data blocks (peach boxes) showing the information passed between functions.}
    \label{fig:proposed_alg}
\end{figure}
In this section, we introduce our proposed DRL-based approach for the rack movement optimization problem. As outlined in Section \ref{sec:problem_formulation}, the problem's complexity scales with the number of positions, rack types, scopes, and resource types, making real-time optimization challenging. To address this, we decompose the original problem into subproblems by rack type in Section \ref{subsec: subproblem} and employ a gradient-based heuristic to sequentially solve each subproblem in Section \ref{subsec: gradient_based_heuristic}. The order of these suboptimizations is important to achieving optimal system performance.

To determine the optimal sequence for suboptimizations, we leverage Leader Reward \cite{wang2024leaderreward}, a method proven effective for NP-hard combinatorial problems such as the TSP and CVRP. We further adapt Leader Reward to our context by encoding each rack type’s attributes—demand, current quantity, resource types, and rack groups—into two float values. These values are then converted into embeddings by the encoder, allowing the decoder to learn the optimal sequence for rack type suboptimizations.

During training, the decoder generates permutations of rack types, representing different suboptimization sequences. The new rack-position mapping is then obtained by the greedy heuristic in Algorithm \ref{alg: gradient_based_heuristic} given the permutations (line 4 in Algorithm \ref{alg: hirarchical_drl_alg}). Each permutation’s performance is evaluated using cumulative rewards, which reflect objective in \eqref{eq: gradient_objective} with the new rack-position mapping as input. By leveraging multiple trajectories and a shared reward baseline, the proposed framework effectively explores the solution space, learning an optimal policy for suboptimization sequencing.

In summary, our algorithm combines a gradient-based heuristic for rack-position mapping with a DRL-based solver to determine the optimal suboptimization order, as outlined in Algorithm \ref{alg: hirarchical_drl_alg}.

\begin{algorithm}[htbp]
\caption{Adapted Leader Reward Training} \label{alg: hirarchical_drl_alg}
\SetAlgoLined
\KwIn{Number of training epochs $T$, batch size $B$, problem set $P$, number of rack types $\mathbf{K}$}
\KwOut{Updated policy network parameter $\theta$}

Initialize policy network parameter $\theta$\;
\For{$\text{epoch} = 1$ \KwTo $T$}{
    $\delta_i \gets \text{GENERATEPROBLEMS}(P), \forall i \in \{1, \dots, B\}$ where $\delta_i= (\bar{X}, d_1, ..., d_{\mathbf{K}}, q, (a_1, b_1), ..., (a_{\mathbf{K}}, b_{\mathbf{K}}))_i$ with $\bar{X}$: initial rack-position mapping and $(a_k, b_k)$: Inputs for the encoder derived from rack type attributes for each $k\in\mathbf{K}$ \;
    $\tau_i^j, X_i^j \gets \text{ROLLOUT+GRADIENTHEURISTIC}(j, \delta_i, \pi_\theta), \forall i \in \{1, \dots, B\}, \forall j \in \{1, \dots, \mathbf{K}\}$\;
    $b_i \gets \frac{1}{N} \sum_{j=1}^{N} Rew(X_i^j, \delta_i), \forall i \in \{1, \dots, B\}$\;
    $\nabla_\theta J(\theta) \gets \frac{1}{BN} \sum_{i=1}^{B} \sum_{j=1}^{N} (Rew(X_i^j, \delta_i) - b_i) \nabla_\theta \log p_\theta(\tau_i^j)$\;
    $\theta \gets \theta + \alpha \nabla_\theta J(\theta)$\;
}
\end{algorithm}

%% file: 6simulation.tex
\section{Simulation}
In this section, we evaluate the performance of our proposed algorithm. We begin with the system setup in Section \ref{subsec: system_setup}, clarify the baselines in Section \ref{subsec: baselines}, and show the performance comparison in Section \ref{subsec: main_results}.
\subsection{System Setup}\label{subsec: system_setup}

We simulate the hierarchical data centers by three sets of position scopes, $\mathcal{S}_1$, $\mathcal{S}_2$, and $\mathcal{S}_3$, representing data centers, suites, and MSBs. Each scope is a subset of $\mathbf{P}$ and any scope in $\mathcal{S}_i$ is contained in one of the scopes in $\mathcal{S}_j$ for $(i,j)\in\{(3,2), (2,1)\}$. Scopes within each set are equal in their size and exclusive to each other. The union of scopes in each set is $\mathbf{P}$ and the union of $\mathcal{S}_1$, $\mathcal{S}_2$, and $\mathcal{S}_3$ is $\mathcal{S}$ in \eqref{eq: resource_limit_constr}. The resource limit matrices for sets $\mathcal{S}_1$, $\mathcal{S}_2$, and $\mathcal{S}_3$ are denoted as $L_1$, $L_2$, and $L_3$, respectively. Concatenation of $L_1$, $L_2$, and $L_3$ along the row dimensions is equal to $L$ in \eqref{eq: resource_limit_constr}.
Table \ref{table: 1} delineates the key parameters considered in this work. The counts of positions, rack types, resource types are symbolized as $|\mathbf{P}|$, $|\mathbf{K}|$, $|\mathbf{R}|$, $|\mathcal{S}|$, $d$, and, $q$, respectively. Each scope in $\mathcal{S}_1$, $\mathcal{S}_2$, $\mathcal{S}_3$ has $500$, $100$, $20$ positions, respectively. The demands for each rack types are uniformly sampled in the range $[d_{min}, d_{max}]$ and placement limit is sampled uniformly from the range $[q_{min}, q_{max}]$. Finally, we use the standard deviation and softplus functions for $\tau(\cdot)$ and $\zeta(\cdot)$, respectively, in our implementation to measure resource spread and penalize resource limit violations, as formulated in \eqref{eq: resource_spread} and \eqref{eq: resource_limit_constr}, making the problem in \eqref{eq:optimization_problem} a INLP.
This datacenter setup is designed by domain experts among the authors, approximating a real-life scenario. 

The matrix $R$ representing the rack-type-to-resource-type mapping is shown in Table \ref{table: 2}.
The previous rack-position mapping is generated using the probability distribution in Table \ref{table: 3}, which has been reviewed and approved by a domain expert among the authors to reflect a realistic scenario. This distribution shows that the no-rack category ($\phi$) has the highest probability at 0.5, meaning that 50\% of positions are unassigned. Among rack types, rack type 7 has the highest probability at 0.204, making it the most likely to be assigned when a rack is present, followed by rack types 5 and 9, with probabilities of 0.102 and 0.051, respectively. The remaining rack types have lower probabilities, ranging from 0.009 to 0.029.

Our proposed algorithm is trained using a single CPU core and a single GPU core on a system with two Milan CPUs @ 2.45GHz and four NVIDIA A100 GPUs. 

\begin{table}[htbp] 
\small
\caption{Summary of the Rack Type Resources} \label{table: 2}
\centering
 \begin{tabular}{|| c || c | c | c | c | c | c | c | c | c | c ||} 
 \hline
 Resource Type & 0 & 1 & 2 & 3 & 4 & 5 & 6 & 7 & 8 & 9 \\
 \hline
 Rack Type 0 & 0 & 0 & 0 & 0 & 0 & 1 & 1 & 0 & 1 & 0 \\
 \hline
 Rack Type 1 & 0 & 1 & 0 & 0 & 1 & 1 & 0 & 0 & 0 & 0 \\
 \hline
 Rack Type 2 & 0 & 0 & 0 & 0 & 1 & 0 & 1 & 0 & 1 & 0 \\
 \hline
 Rack Type 3 & 0 & 1 & 1 & 0 & 0 & 0 & 1 & 0 & 0 & 1 \\
 \hline
 Rack Type 4 & 0 & 1 & 0 & 1 & 0 & 0 & 0 & 0 & 1 & 0 \\
 \hline
 Rack Type 5 & 0 & 0 & 1 & 1 & 1 & 0 & 1 & 0 & 1 & 0 \\
 \hline
 Rack Type 6 & 1 & 0 & 1 & 1 & 0 & 0 & 0 & 0 & 0 & 0 \\
 \hline
 Rack Type 7 & 1 & 1 & 0 & 1 & 0 & 0 & 1 & 0 & 0 & 0 \\
 \hline
 Rack Type 8 & 0 & 0 & 0 & 0 & 1 & 0 & 0 & 0 & 1 & 0 \\
 \hline
 Rack Type 9 & 1 & 1 & 0 & 0 & 0 & 0 & 0 & 0 & 0 & 1 \\
 \hline
 \end{tabular}
\end{table}

\begin{table}[bthp] 
\caption{Probability Distribution of Rack Types for Each Position} \label{table: 3}
\centering
 \resizebox{\columnwidth}{!}{%
\begin{tabular}{|| c | c | c | c | c | c | c | c | c | c | c | c ||} 
 \hline
 Rack Type & 0 & 1 & 2 & 3 & 4 & 5 & 6 & 7 & 8 & 9 & $\phi$\\
 \hline\hline
Probability & 0.033 & 0.014 & 0.030 & 0.010 & 0.009 & 0.102 & 0.029 & 0.204 & 0.018 & 0.051 & 0.5\\
 \hline
 \end{tabular}
 }
\end{table}

\begin{figure*}[htbp]
    \centering
    \begin{subfigure}[b]{0.32\textwidth}
        \centering
    \includegraphics[width=\textwidth]{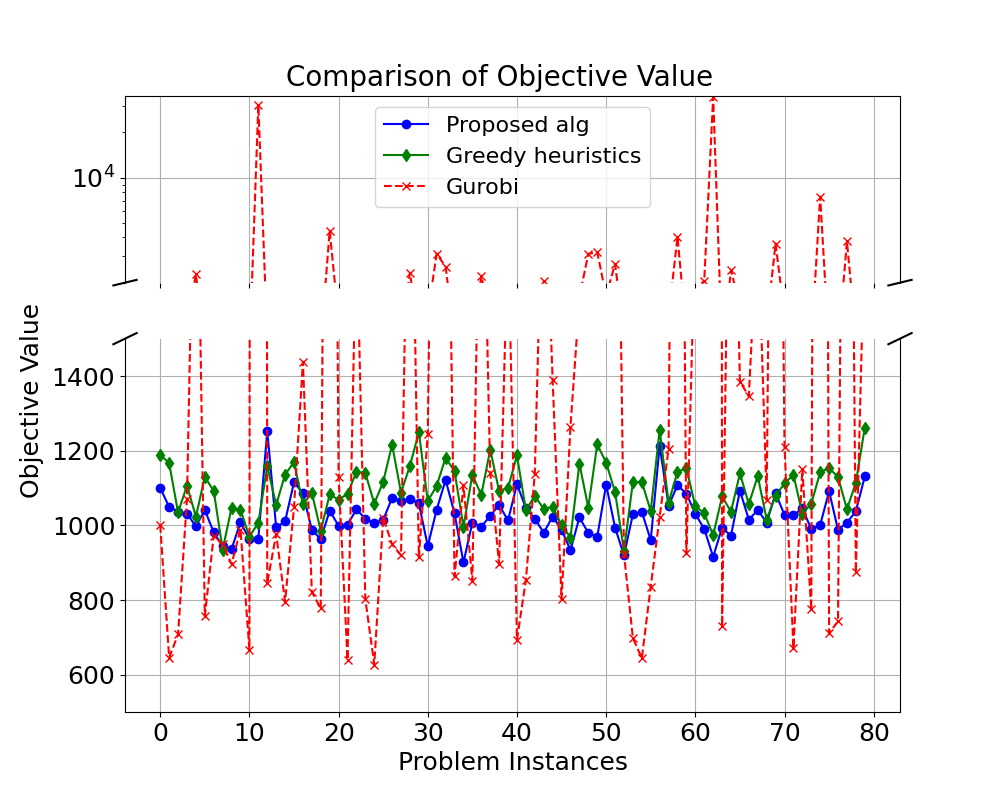}
        \caption{Comparison of Objective Value}
        \label{fig:obj}
    \end{subfigure}
    \begin{subfigure}[b]{0.32\textwidth}
        \centering
        \includegraphics[width=\textwidth]{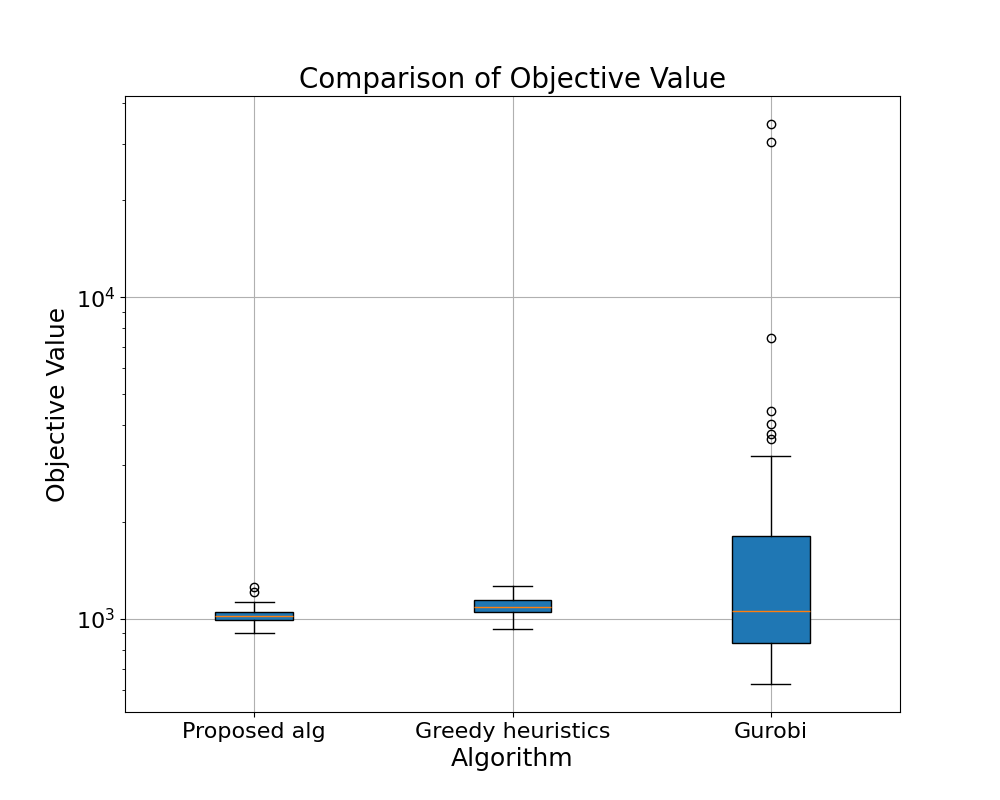}
        \caption{Objective Value Statistics }
        \label{fig:obj_bp}
    \end{subfigure}
    \begin{subfigure}[b]{0.32\textwidth}
        \centering
        \includegraphics[width=\textwidth]{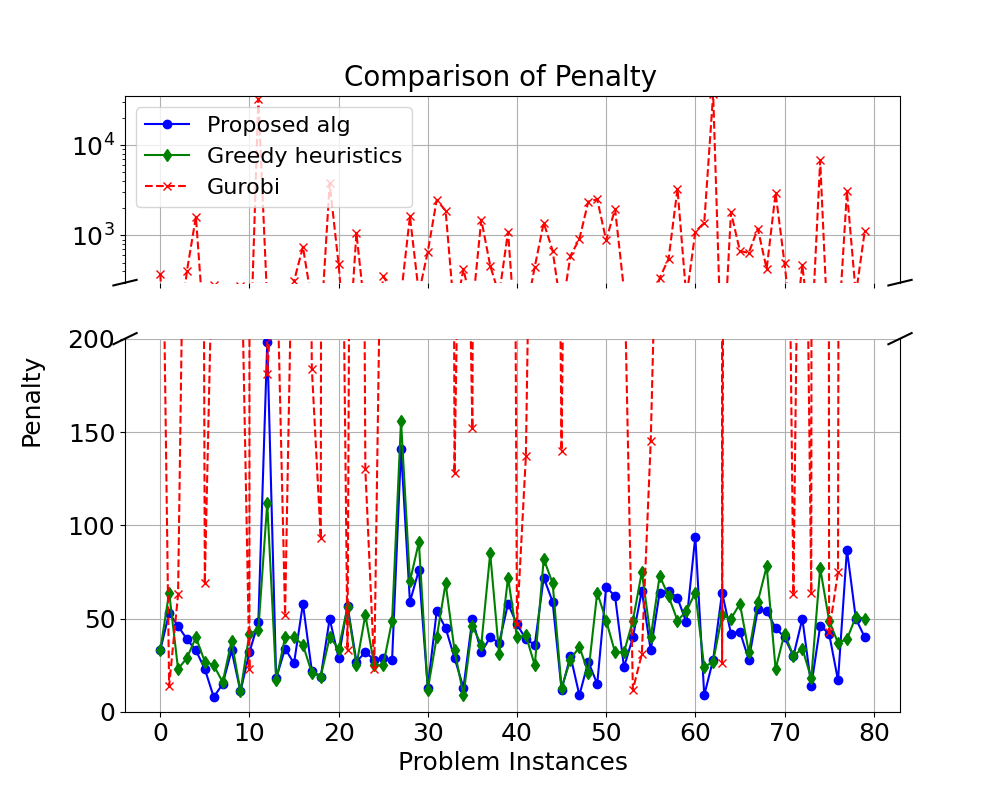}
        \caption{Comparison of Penalty}
        \label{fig:penalty}
    \end{subfigure}
    \begin{subfigure}[b]{0.32\textwidth}
        \centering
    \includegraphics[width=\textwidth]{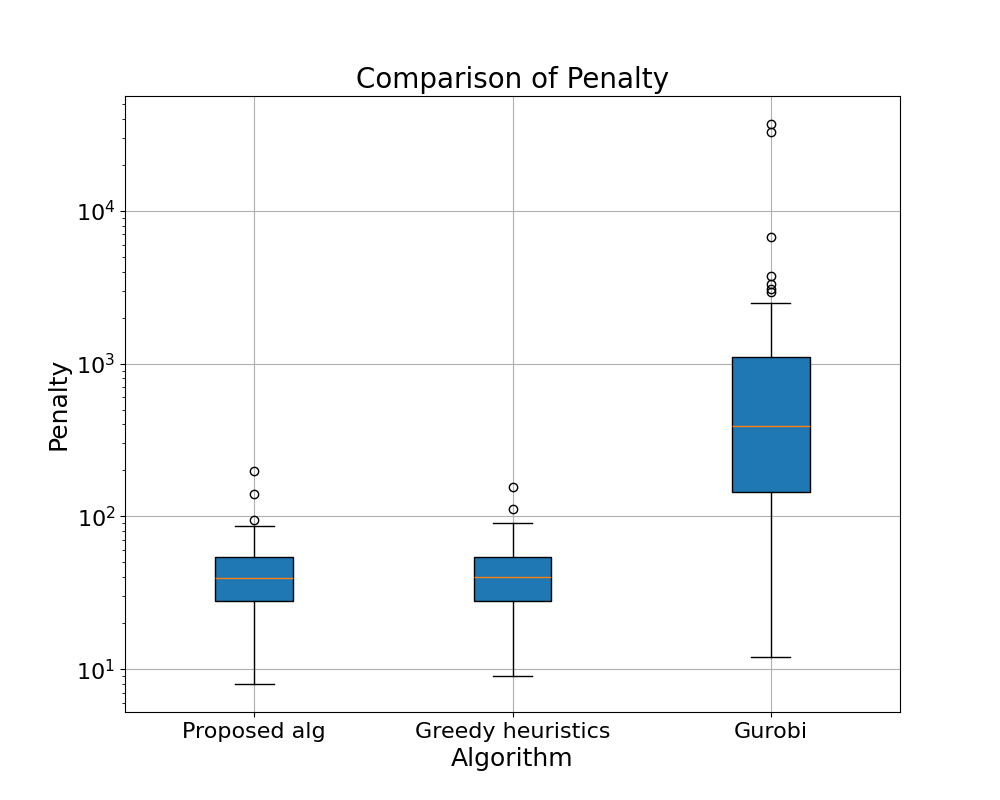}
        \caption{Comparison of Penalty Statistics}
        \label{fig:penalty_bp}
    \end{subfigure}
    \begin{subfigure}[b]{0.32\textwidth}
        \centering
        \includegraphics[width=\textwidth]{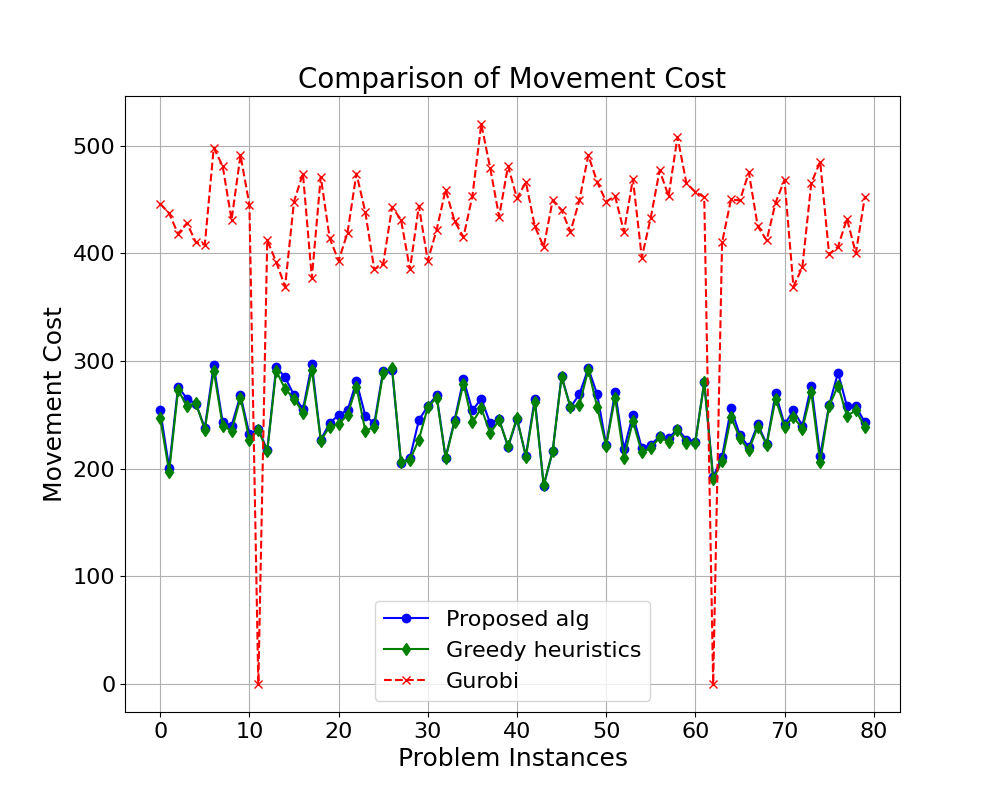}
        \caption{Comparison of Movement Count}
        \label{fig:movement_cost}
    \end{subfigure}
    \begin{subfigure}[b]{0.32\textwidth}
        \centering
        \includegraphics[width=\textwidth]{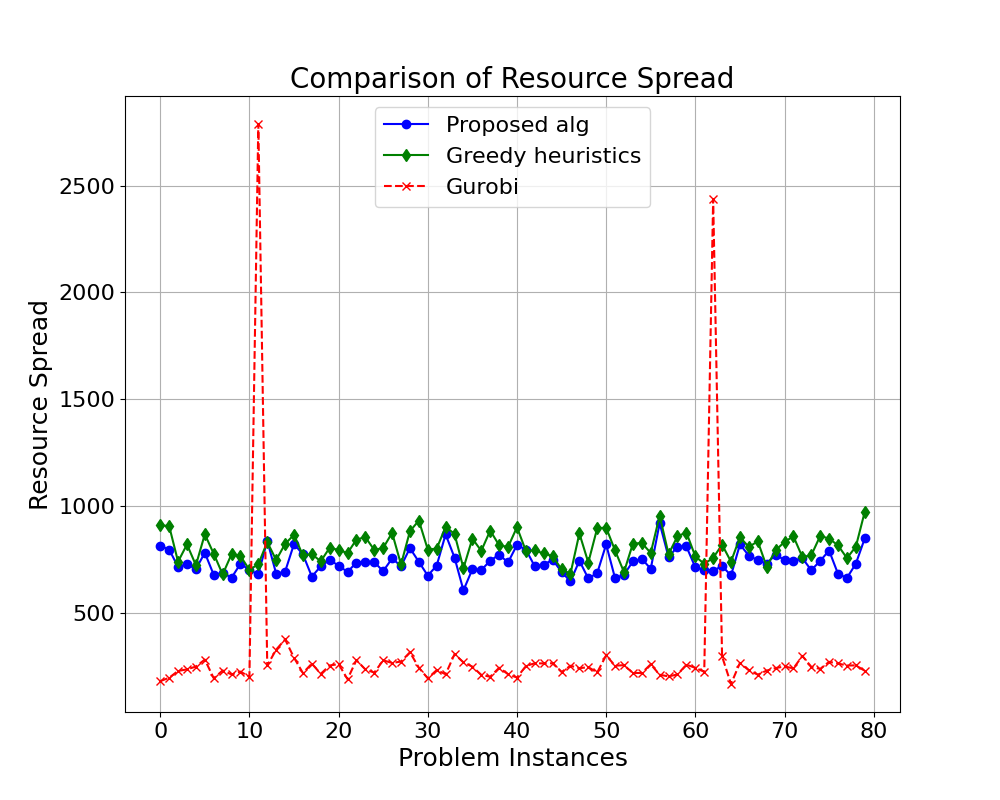}
        \caption{Comparison of Resource Spread}
        \label{fig:std}
    \end{subfigure}

    \caption{Comparison of the Proposed Algorithm, gradient Heuristics, and Gurobi across different metrics.}
    \label{fig:comparison}
\end{figure*}
\subsection{Baseline Algorithms} \label{subsec: baselines}

This section describes the baseline algorithms. We include the MIP solver, which was widely used for operation optimizations, and the gradient-based heuristic proposed in this paper. The descriptions of the baseline algorithms are given as follows. 
\begin{itemize}
    \item MIP solver: We implemented the solver using Gurobi v11.0.3 \cite{gurobi}. The solver's time limit configuration and performance analysis are presented in the Section \ref{subsec: main_results}. Gurobi was run using 18 threads on an Intel(R) Core(TM) i9-9980XE CPU @ 3.00GHz. When Gurobi cannot find a solution within the specified time limit, we retain the previous rack-type-to-position mapping as the solution.
    \item Gradient-based heuristics: This is implemented by PyTorch \cite{PyTorch} following Algorithm \ref{alg: gradient_based_heuristic} to allocate positions for one rack type at a time. The order of rack is predefined.
\end{itemize}

\begin{table*}[htbp] 
\small
\caption{Summary of the algorithm capability} \label{table: 4}
\centering
 \begin{tabular}{|| c || c | c | c | c | c ||} 
 \hline
 (Algorithm, Time limit (min) )& (MIP, 1) & (MIP, 5) & (MIP, 10) & (MIP, 20) &(proposed, N/A) \\
 \hline\hline
Solving percentage (\%) & 0 & 6.25  & 81.25 & 97.5 & 100\\
 \hline
 \makecell[c]{Running time (min)} & 110 & 630 &830  & 1630 & 2\\
 \hline
 \makecell[c]{Training time (min)} & N/A & N/A & N/A  & N/A & 600\\
 \hline
 \makecell[c]{Performance $\mathcal{U}$} &38816 & 1600 &1563 &1475 & 1025 \\
 \hline
 \end{tabular}
\end{table*}
\begin{figure*}[htbp]
    \centering
    \begin{subfigure}{0.32 \textwidth}
        \includegraphics[width=\textwidth]{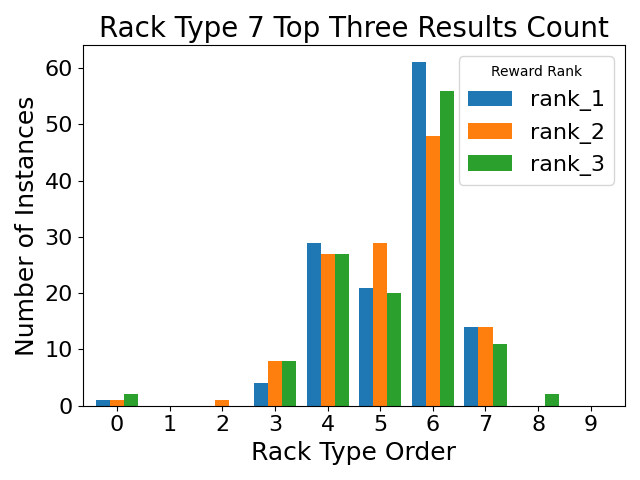}
        \caption{}
        \label{fig:rt7}
    \end{subfigure}
    \begin{subfigure}{0.32 \textwidth}
        \includegraphics[width=\textwidth]{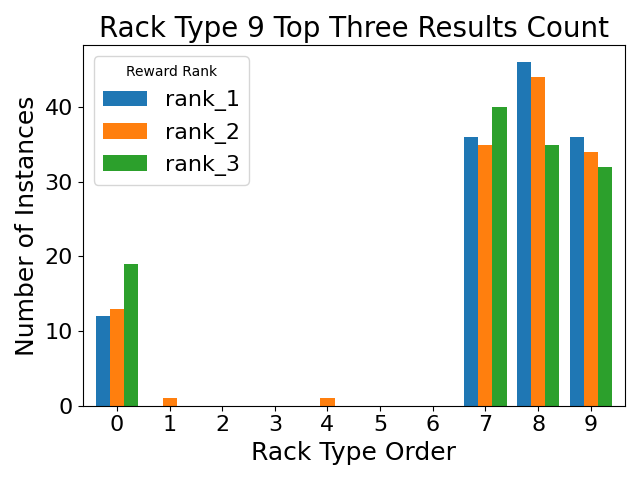}
        \caption{}
        \label{fig:rt9}
    \end{subfigure}
    \begin{subfigure}{0.32 \textwidth}
        \includegraphics[width=\textwidth]{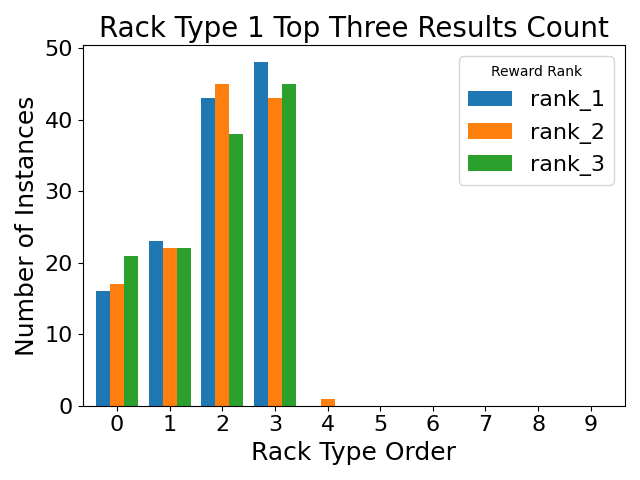}
        \caption{}
        \label{fig:rt1}
    \end{subfigure}
    \caption{Top-three reward rankings for Rack Type 1, Rack Type 7, and Rack Type 9 across 80 $\times$ 10 problem instances. Each figure shows the number of instances where the respective rack type, placed in a specific position within the order, achieved a top-three reward ranking (rank\_1, rank\_2, or rank\_3). The x-axis represents the order position of the rack type, while the y-axis shows the number of instances. Rack Type 1 performs best when placed in position 3, Rack Type 7 performs optimally in position 6, and Rack Type 9 achieves the highest rewards when placed in position 8, indicating preferred order positions for each rack type to maximize performance.}
    \label{fig:rt_order}
\end{figure*}

\subsection{Main Results} \label{subsec: main_results}
We compare the performance of our proposed algorithms with the baseline algorithms in this section.  In Figure \ref{fig:comparison}, we compare the baselines and our algorithm on the overall performance, resulting in movement count, resource limit violation penalty, and resource spread metric. 
We compare the execution time of MIP and our proposed algorithm in Table \ref{table: 4}. Finally, we discuss the decision made by Leader Reward given attributes of the rack types in Figure \ref{fig:rt_order}.

Figure \ref{fig:comparison} compares the performance of three algorithms — our proposed algorithm, gradient-based heuristics, and MIP solver — across 80 problem instances. The comparison is based on four key metrics: movement count, objective value, resource limit penalty, and resource spread metric, where the objective value is the summation of the other three metrics. 
In Figure \ref{fig:obj}, the proposed algorithm and gradient-based heuristics show similar performance, with objective values ranging between 1000 and 1400. Moreover, the proposed algorithm outperforms the gradient-based heuristics for nearly all of the 80 problem instances and the average objective values of the proposed algorithm outperforms the gradient-based heuristics by 7\%. Comparing with that the results for the other two algorithms remain consistent and relatively smooth across all problem instances, Gurobi presents a more erratic behavior. Although it maintains reasonable objective values for most instances, there are extreme spikes where the objective value exceeds $10^4$, where the solver reaches the time limit of 20 minutes and cannot get a solution. Among the 80 problem instances, the proposed algorithm outperforms the MIP solver for 46 instances and the average objective value of the proposed algorithm outperforms the one for the MIP solver by more than 30\%. Figure \ref{fig:movement_cost}, \ref{fig:penalty}, and \ref{fig:std} show that the MIP solver succeeds in minimizing resource spread metrics, but incurs high movement count and resource limit penalty. At instances 12 and 63, where the MIP solver cannot get a solution in time, the previous rack-type-to-position mapping is used as the action and thus incurs zero movement count and high penalty and resource spread. In conclusion, the proposed algorithm and gradient-based heuristic demonstrate minimal variability, highlighting their robustness in addressing problem instances—an essential feature for large-scale data center management. In contrast, the MIP solver demonstrates significant instability, with sharp spikes in all metrics for certain instances. This suggests that while the MIP solver may occasionally achieve competitive results, it struggles to maintain consistent performance, particularly in terms of balancing the components of the objective function.

Table \ref{table: 4} presents a comparative analysis of the MIP solver with various time limits (1, 5, 10, and 20 minutes) and the proposed algorithm, evaluated across multiple metrics: solving percentage, running time, training time, and objective value.
The proposed algorithm demonstrates superior performance in all metrics. It successfully solves 100\% of the problem instances, outperforming all MIP configurations, even those with extended time limits. In contrast, MIP with shorter time limits (1 and 5 minutes) struggles, solving 0\% and 6.25\% of the instances, respectively. MIP shows improved performance with longer time limits—solving 81.25\% and 97.5\% of instances at 10 and 20 minutes, respectively—though it still falls short of the proposed method.
Regarding running time, the proposed approach is remarkably efficient, requiring just 2 minutes to complete all problem instances. By comparison, MIP algorithms require significantly more time, with running times ranging from 110 minutes for the 1-minute time limit to 1630 minutes for the 20-minute time limit. This stark contrast in computational efficiency highlights the practicality of the proposed method.
While training time is not applicable to MIP, the proposed algorithm requires 600 minutes of training time. However, this one-time training time is offset by its exceptional runtime efficiency.
In terms of objective value, the proposed algorithm achieves the best performance with a score of 1025, outperforming all MIP variants. MIP’s performance improves as the time limit increases, with the best MIP configuration (20 minutes) yielding a performance of 1475. However, even this best-case MIP performance lags behind the proposed algorithm. In Appendix \ref{apdx: scalabilty}, we provide additional results analyzing how the training time scales with the number of rack types, further illustrating the efficiency of our proposed algorithm.

Fig. \ref{fig:rt_order} illustrate the performance of different rack type orders for three specific rack types - Rack Type 1, Rack Type 7, and Rack Type 9 - evaluated across 80 $\times$ 10 problem instances using a Leader Reward agent. As shown in Table \ref{table: 3}, Rack Type 1, Rack Type 7, and Rack Type 9 are selected for the plots because they are either very unlikely or likely to be found in the data center positions previously. Each figure shows the number of instances where the respective rack type, placed in a specific order position, achieved a top-three reward ranking (rank\_1, rank\_2, or rank\_3). The objective is to identify the optimal position for each rack type within the order that maximizes its reward. For Rack Type 1, the optimal position is found to be position 3, where it achieves the highest number of top-three rewards across the problem instances. Positions beyond 4, however, show a sharp decline in performance, indicating that Rack Type 1 is most effective when placed early in the sequence, particularly around the third position. The performance of Rack Type 7 is clearly optimized when it is placed in position 6. Positions 4 and 5 also perform well, although to a lesser degree, suggesting that Rack Type 7 benefits from being placed toward the latter half of the order. For Rack Type 9, the ideal position is position 8, which consistently delivers the highest number of top-three rewards across the instances. Positions 7 and 9 also contribute to a significant number of top-three results, particularly in rank\_2 and rank\_3, indicating that Rack Type 9 performs best when placed at or near the end of the sequence.

%% file: 7conclusion.tex
\section{Conclusions}

This work presents a scalable, DRL-based solution for optimizing rack movement in large, heterogeneous data centers. Our two-tier approach, combining Leader Reward with a greedy heuristic, outperforms traditional MIP solvers and heuristics by reducing movement counts, ensuring fault tolerance, and improving execution times. Simulations confirm that our method achieves efficient, consistent results across varying demands, highlighting its practical application in dynamic data center environments. 

\section{Acknowledgment}

C. Chen, J. Chen, T. Lan, and V. Aggarwal acknowledge research award from Meta Platforms, Inc., and Office of Naval Research under grant N00014-23-1-2532.

%% file: apdx.tex
\begin{figure*}[t]
    \centering
    \begin{subfigure}{0.4 \textwidth}
        \includegraphics[width=\textwidth]{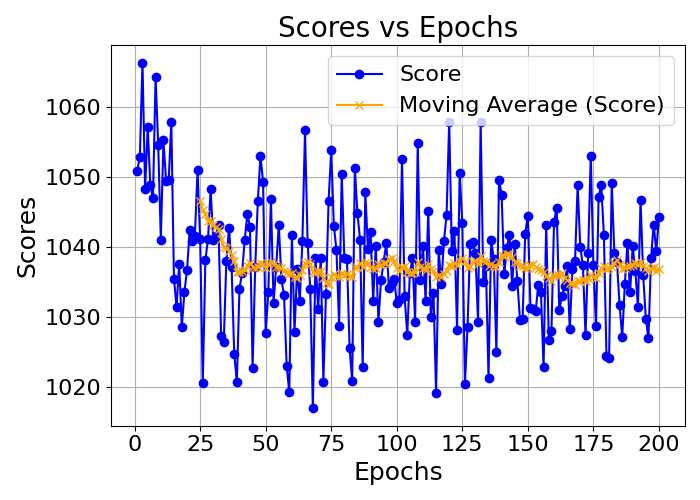}
        \caption{}
        \label{fig:conv_obj}
    \end{subfigure}
    \begin{subfigure}{0.4 \textwidth}
        \includegraphics[width=\textwidth]{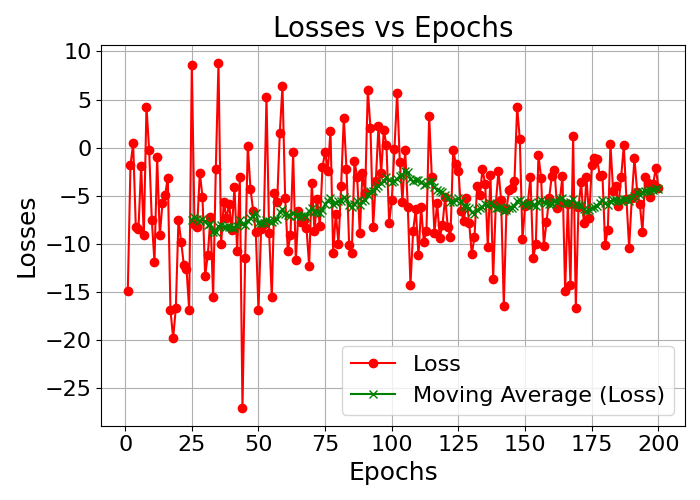}
        \caption{}
        \label{fig:conv_loss}
    \end{subfigure}
    \caption{(a) shows the agent’s scores across epochs, with the blue line representing raw scores and the orange line indicating the moving average of the scores over a 25-epoch window. (b) displays the loss values over epochs, where the red line corresponds to raw losses, and the green line represents the moving average of losses. Both metrics illustrate the agent’s learning process, characterized by high variability during early epochs followed by gradual stabilization in later epochs.}
    \label{fig:conv}
\end{figure*}

\section{Notation summary} \label{apdx: notation_summary}
\begin{table}[htbp] 
\caption{Summary of Notations} \label{table: notation_summary}
\centering
\small  
\setlength{\tabcolsep}{4pt}  
 \begin{tabular}{|c|l|} 
 \hline
 \textbf{Notation} & \textbf{Description}  \\ 
 \hline
 $\mathbf{P}$ & Set of positions \\ 
 \hline
 $\mathbf{K}$ & Set of rack types \\ 
 \hline
 $\bar{\mathbf{K}}$ & Set of optimized rack types in heuristics \\ 
 \hline
 $\mathbf{R}$ & Set of resource types \\ 
 \hline
 $\mathcal{S}$ & Set of scopes \\ 
 \hline
 $\mathcal{G}$ & Set of subsets of $\mathbf{K}$ \\ 
 \hline
 $R$ & The rack-type-to-resource-type binary matrix  \\
 \hline
 $S$ & The position-to-scope binary matrix  \\
 \hline
 $X$ & The position-to-rack-type binary matrix  \\
 \hline
 $G$ & A subset of $\mathbf{K}$  \\
 \hline
 $L$ & The resource limit matrix  \\
 \hline
 $d_k$ & Placement request for rack type $k$ \\ 
 \hline
 $q$ & Total placement limit \\ 
 \hline
 $M_k$ & Movement weight for rack type $k$ \\ 
 \hline
 $\tau(\cdot)$ & Resource spread function \\ 
 \hline
 $\zeta(\cdot)$ & Resource limit violation penalty function \\ 
 \hline
 $I$ & Set of resource spread requirements\\
 \hline
 $\beta_1$ & \makecell[l]{Weights for resource spread metric in\\ the objective function}\\
 \hline
 $\beta_2$ & \makecell[l]{Weights for resource limit violation in\\ the objective function}\\
 \hline
 $\gamma$ & \makecell[l]{Weights for placement limit violation in\\ the augmented objective function in heuristic}\\
 \hline
 $\Gamma$ & Number of rack adjustment in heuristic \\
 \hline
 $T$ & Number of training epochs\\
 \hline
 $B$ & Batch size\\
 \hline
 $\mathcal{P}$ & Problem set for training\\
 \hline
 \end{tabular}
\end{table}

\section{Discussion}


This paper presented a novel approach for optimizing rack-position mapping in rack-level composable, heterogeneous data centers. We formulated the problem as an integer non-linear optimization problem, aiming to minimize a weighted combination of rack movement counts and resource distribution metrics while respecting the resource constraints within the data center. To solve this large-scale optimization problem, which involves more than 100,000 positions and approximately 100 rack types, we proposed a scalable solution using a deep reinforcement learning (DRL)-based framework that incorporates Leader Reward and a gradient-based heuristic.
Simulation results demonstrate the effectiveness of our proposed algorithm in comparison with both a gradient-based heuristic and a Mixed Integer Programming (MIP) solver. Our method outperformed the gradient-based heuristic, achieving a 7\% improvement in performance on average, while also outperforming the MIP solver by over 30\% in average objective value across 80 problem instances. Furthermore, the proposed algorithm achieved a 100\% problem-solving success rate, compared to the MIP solver's 97.5\% at its maximum time limit of 1200 seconds, and demonstrated significant computational efficiency with a runtime of just 2 minutes, compared to MIP's 1630 minutes.
In contrast to the MIP solver, which exhibited instability and significant variability in performance due to time limits and high resource penalties, our algorithm provided robust and consistent performance across all problem instances. The MIP solver’s tendency to incur high movement counts and resource penalties, particularly when unable to complete computations within the time limit, underscores the limitations of traditional optimization methods for large-scale, heterogeneous cloud systems.
In conclusion, the proposed DRL-based solution offers a highly efficient and scalable approach for managing rack placement and adjustment in complex data centers. It achieves superior performance in terms of solution quality, computational efficiency, and consistency compared to both traditional MIP solvers and heuristic methods. These results suggest that the proposed method is well-suited for practical deployment in large-scale cloud systems, where rapid changes in hardware and resource demands require robust and flexible optimization strategies.

%% file: apdx_sims.tex
\section{Convergence of the Proposed Approach}


In this section, we show the convergence  in terms of the number of epochs required to reach convergence. Figure \ref{fig:conv} displays the training dynamics of the Leader Reward agent over $200$ epochs, with Figure \ref{fig:conv_obj} depicting the evolution of objective values and Figure \ref{fig:conv_loss} illustrating the changes in the loss function. Figure \ref{fig:conv} illustrates that while the Leader Reward agent experiences considerable fluctuations in both scores and losses during the early stages of training, there is a noticeable stabilization in performance after approximately 50-75 epochs. The moving averages of both scores and losses suggest that the agent learns a reasonably effective policy early on, which may be sufficient to meet performance requirements. However, the continued presence of variability, particularly in later epochs, indicates that the policy has not fully converged, and extended training may be necessary to further optimize the agent’s performance and ensure more consistent results.

\section{Scalability of Training Time} \label{apdx: scalabilty}
Table \ref{table: 4} in the manuscript demonstrates that the total time (training + inference) for our proposed algorithm is significantly lower than Gurobi’s runtime, making it a more practical solution. Moreover, as training is conducted offline, it does not affect real-time decision-making.

To further illustrate how training time scales with system complexity, we present additional results for different numbers of rack types in the table below.

\begin{table}[htbp] 
\small
\caption{Scalability of Training Time} \label{table: 5}
\centering
 \begin{tabular}{|| c || c | c | c | c | c ||} 
 \hline
 Number of rack types& 10 & 15 & 20 & 30 & 35\\
 \hline
Training time (min) & 600 & 1200  & 1800 & 4500 & 6000\\
\hline
 \end{tabular}
\end{table}